\documentclass[lettersize,journal]{IEEEtran}

\usepackage{cite}

\usepackage{amsmath,amssymb,amsfonts}
\usepackage{graphicx}
\usepackage{float}
\usepackage{textcomp}
\usepackage{xcolor}
\usepackage{diagbox}
\usepackage{graphics}
\usepackage[utf8]{inputenc}
\usepackage{booktabs} 
\usepackage[caption=false,font=normalsize,labelfont=sf,textfont=sf]{subfig}
\usepackage{subcaption}
\usepackage{xspace}
\usepackage{makecell}
\usepackage{multicol}
\usepackage{multirow}
\usepackage{tikz}
\usepackage{arydshln}
\usepackage{bm}
\usepackage{algorithm}
\usepackage{algorithmic}
\usepackage{amsmath}
\usepackage{paralist}
\usepackage{url}
\usepackage{threeparttable}
\newcommand{\sysname}[0]{{EdgeOAR}\xspace}

\begin{document}
\title{\sysname: Real-time Online Action Recognition On Edge Devices}

\author{Wei~Luo,~\IEEEmembership{Student~Member,~IEEE},~Deyu~Zhang,~\IEEEmembership{Member,~IEEE}, ~Ying~Tang~\IEEEmembership{Student~Member,~IEEE}, 
~Fan~Wu,~\IEEEmembership{Member,~IEEE}, ~Yaoxue~Zhang,~\IEEEmembership{Senior Member,~IEEE}
	\IEEEcompsocitemizethanks{
		\IEEEcompsocthanksitem Wei Luo, Deyu Zhang and Fan Wu are with the School of Computer Science and Engineering, Central South University. Changsha 410083, China. E-mails:  \{weiluo, zdy876, wfwufan\}@csu.edu.cn.
        \IEEEcompsocthanksitem Ying Tang is with the Big Data Institute, Central South University. Changsha 410083, China. E-mail: tangyin0512@csu.edu.cn.
		\IEEEcompsocthanksitem Yaoxue~Zhang is with the Department of Computer Science and Technology, Tsinghua University, Beijing 100084, China. E-mail: zhangyx@mail.tsinghua.edu.cn. 
        \IEEEcompsocthanksitem Deyu Zhang is the corresponding author.
	}

}

\markboth{Journal of \LaTeX\ Class Files,~Vol.~14, No.~8, August~2021}%
{Shell \MakeLowercase{\textit{et al.}}: A Sample Article Using IEEEtran.cls for IEEE Journals}


\maketitle
\begin{abstract}
This paper addresses the challenges of Online Action Recognition (OAR), a framework that involves instantaneous analysis and classification of behaviors in video streams. OAR must operate under stringent latency constraints, making it an indispensable component for real-time feedback for edge computing. Existing methods, which typically rely on the processing of entire video clips, fall short in scenarios requiring immediate recognition. To address this, we designed \sysname, a novel framework specifically designed for OAR on edge devices. 
\sysname includes the Early Exit-oriented Task-specific Feature Enhancement Module (TFEM), which comprises lightweight submodules to optimize features in both temporal and spatial dimensions. We design an iterative training method to enable TFEM learning features from the beginning of the video. Additionally, \sysname includes an Inverse Information Entropy (IIE) and Modality Consistency (MC)-driven fusion module to fuse features and make better exit decisions. This design overcomes the two main challenges: robust modeling of spatio-temporal action representations with limited initial frames in online video streams and balancing accuracy and efficiency on resource-constrained edge devices.
Experiments show that on the UCF-101 dataset, our method \sysname reduces latency by 99.23\% and energy consumption by 99.28\% compared to state-of-the-art (SOTA) method. And achieves an adequate accuracy on edge devices.



\end{abstract}
\section{Introduction}

Action recognition aims to recognize human activities such as push-ups, running, and walking within a video segment.\cite{simonyan2014two,zhang2016real} This capability serves as a key component for numerous edge applications. For end-users, particularly in the context of Online Action Recognition (OAR), it is crucial to deliver action recognition results from video segments captured by smartphones as quickly as possible. This meets the needs of downstream tasks, especially in scenarios such as emergency response in camera surveillance, real-time feedback in fitness applications, and providing immersive experiences in mobile games. This rapid processing is essential because delays can significantly impact user experience and the effectiveness of the application.

Unfortunately, we identify a significant gap between existing practices and the goal of achieving optimal responsiveness. The current solution typically requires the complete video segments as a prior. 
Most action recognition tasks typically involve uniform sampling across the entire video for action modeling\cite{wang2018temporal,wu2018compressed,lin2019tsm,zheng2023dynamic,wang2023videomae}. Some works improve the sampling process by selecting key frames to enhance recognition accuracy\cite{pan2023action,yang2022sta}, which still require a global perspective. Other works attempt early exit strategies, but limited to frames sampled from the entire video\cite{ghodrati2021frameexit,wang2023tlee}.
In an online scenario, the latency perceived under the existing solutions is the video segment length plus the duration required for video analysis. Typical video segment lengths are a few seconds, leading to inference latency of more than several seconds, which is an unacceptable delay for edge users. Even taking advantage of stronger computing facilities such as cloud servers, despite potential privacy concerns, does not bridge this gap due to the reliance on processing complete video segments.

The root cause of this issue is that existing action recognition frameworks fail to fully exploit the predictive and recognition capabilities of the model. In contrast, human intelligence seamlessly integrates both abilities, enabling the determination of action types from observing only the initial segments of a video sequence, without necessitating the processing of the entire sequence.
Inspired by human perception, in this paper, we propose the \sysname framework. During the inference phase, \sysname acquires action information from the initial frames of a video. 
Once the reliability threshold for action recognition is reached, our framework can directly deliver the recognition result.
\sysname ensures accuracy in edge devices, significantly reducing latency and energy consumption for action recognition tasks.

Inferring action recognition results accurately from only the beginning of videos presents two new challenges. \textit{First, how to robustly model spatial-temporal action representation from online video streams.} 
In online video streams, the lack of future information means that sufficient information to represent actions can only be captured in a limited number of initial frames. 
Given the diversity and complexity of actions, enhancing the model's modeling capabilities while maintaining a certain level of generalization challenges the model's capacity for accurate representation.
\textit{Second, how to strike the balance between accuracy and efficiency on resource-limited edge devices.}
Deploying deep learning models on edge devices encounters resource constraints, particularly in computing power and energy consumption.
Users expect highly responsive and accurate feedback in online applications.
Using complex models to achieve powerful feature extraction can result in excessively long latency per frame, possibly even failing to complete the analysis before the video segment finishes playing, which is counterproductive. Achieving a balance between accuracy and efficiency on resource-limited edge devices presents another significant challenge.

To address these challenges, we propose \sysname, a general framework that enables real-time OAR on edge devices.
Specifically, we have designed the Early Exit-oriented Task-specific Feature Enhancement Module (TFEM) which consists of two lightweight submodules: the Temporal Layering Shift Module (TLSM) and the MBs-guided Spatial Enhancement Module (MSEM). These modules optimize feature representations in the temporal and spatial dimensions, respectively, to enhance the overall performance of TFEM.
To enhance its performance, we have developed an iterative training method that helps TFEM learn features from the beginning of the video. Additionally, we have designed the Inverse Information Entropy (IIE) and Modality Consistency (MC)-driven Fusion Module to fuse features from different modalities in both time and space, further improving accuracy and aiding the model in making better exiting decisions. Our design is edge-friendly and does not introduce excessive computational overhead.

In summary, the main contributions are as follows:
\begin{itemize}
    \item We design \sysname, a first online action recognition framework,  implemented on Android with nearly 3,000 lines of Java\cite{arnold2005java} code (excluding test code). 
    \item Considering the limited computing power and resources of edge devices, we design a two-stage online framework and develop an iterative training approach to enable the backbone to gradually learn the features from the early part of the video. We then incorporate prior information to adaptively fuse inference results from different modalities and automatically exit based on a threshold.
\item To the best of our knowledge, \sysname is the first online action recognition framework specifically designed for edge devices. Experiments show that on the two public datasets, \sysname reduces latency by an average of 97.24\%, power consumption by 26.56\% and decreases accuracy by 3.26\% compared to offline methods.


\end{itemize}

\section{Background and Motivation}
Action recognition demands not only analyzing the content within each frame but also understanding the temporal relationships between frames to identify the overall pattern of the action. 
We first briefly introduced the encoding process of H.265 videos on edge devices and the relationship between the information contained in the videos and the motion content captured. Then, we outlined the motivation for early exit in action recognition and the feasibility of using multi-modal data to enhance action recognition accuracy.

\subsection{Background:} 
In the task of action recognition, motion representation, such as optical flow, serves as an important supplementary input. It provides motion cues at the pixel level, but is computationally complex.
Fortunately, edge devices are equipped with dedicated codec chips that can compress image streams with little overhead. 
We rapidly extract compressed-domain information encoded in H.265 to facilitate online recognition tasks on edge devices.

The compressed domain information of H.265 videos includes the following:
\textit{Macroblocks (MBs)} constitute the basic units of video encoding.
Video frames are segmented into MBs during encoding, with smaller MBs typically used for complex areas. Thus, MBs distribution partially reflects the image's spatial information.
The H.265 encoding standard provides a more granular representation than its predecessors, with macroblock partition sizes from $4 \times 4$ to $64 \times 64$ across 12 levels.
This enables detailed, macroblock-level comparison of differences between the current and adjacent frames.
\textit{Motion vectors(MVs)} indicate the positional offset of a current macroblock relative to its corresponding macroblock in the reference frame. 
They serve as a macroblock-level alternative to pixel-level optical flow.
\textit{Residuals} denote differences in color, texture, and other attributes between the current and predicted macroblocks.
They highlight post-prediction discrepancies, showing where the predicted macroblock differs from the actual content.

\subsection{Motivation:}

The motivation for this study arises from exploring the efficiency-accuracy trade-off in action recognition tasks.
Firstly, we aim to determine the feasible upper limits of latency and accuracy. Ideally, accurate video action recognition can be achieved by examining just one frame appropriately, achieving optimal latency and accuracy. 
To explore this, we perform single-frame tests using ResNet with MaxPooling to manage multi-modal inputs flexibly and minimize sampling method biases in accuracy.
To ensure classification fairness, we selected 561 videos, each approximately 76$\pm$6 frames long, where 6 represents half the GOP length.
We first evaluate each video frame to assess the feasibility of early exit strategies. 
Subsequently, we leverage multiple modalities, including motion vectors and residuals, to enhance action recognition precision.

\begin{figure}[htb]
	\centering \includegraphics[width=1\columnwidth,height=0.6\columnwidth]{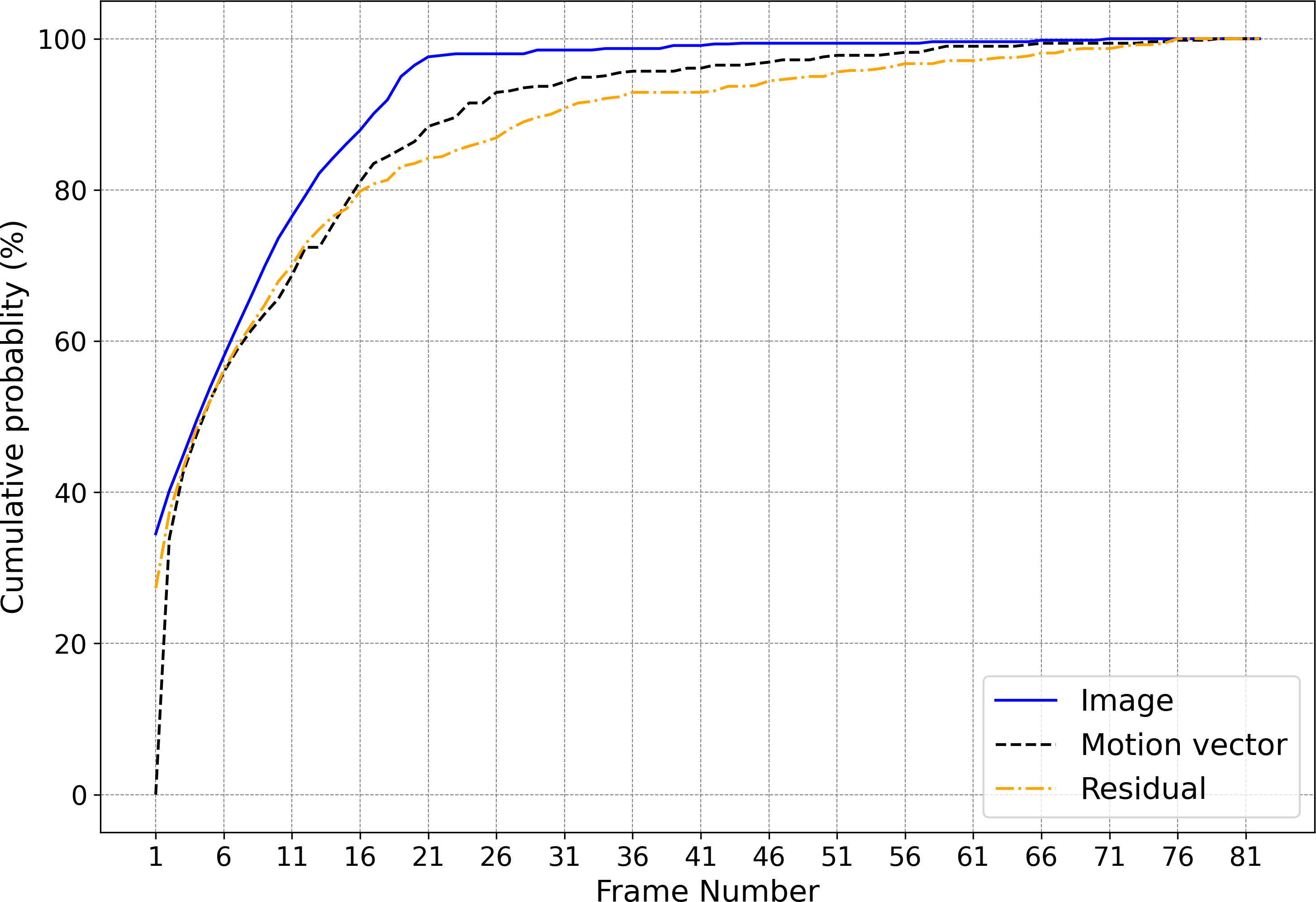}
	\caption{The position of the earliest frame capable of producing correct results.
 }
	\label{F.accuracyVSframenum}
\end{figure}

\textbf{Early frames in video streams significantly contribute to action recognition.} We record the frame numbers at which correct inferences first appeared for each video, as shown in Figure \ref{F.accuracyVSframenum}. 
Observations show that 52.84\% of videos are correctly classified from the first frame in image-based inference.
The average probability of correct classification increases to 77.18\% within the first 10 frames and to 90.54\% within the first 20 frames, which is just 36.14\% of the video duration.
This implies that a sufficiently powerful model can save 87.95\% of time with a 77.18\% chance and 75.90\% of time with a 90.54\% chance by capturing the right frame for inference.
These results indicate that correct classification frames tend to appear early in most video sequences.
Accurately identifying these frames can significantly accelerate the OAR task. 
However, accurately capturing frames to enhance classification accuracy and timely exiting after sufficient frames are captured remain pressing challenges.

\begin{table}[hbt]
	\setlength{\abovecaptionskip}{0pt}
\setlength{\belowcaptionskip}{1pt}
\renewcommand\arraystretch{1.1} 
	\caption{Inference results using different modalities of data.}
	\label{T.multi-modalDataResults}
	\centering
	\begin{threeparttable}
		\setlength{\tabcolsep}{1.8mm}{
			\begin{tabular}{cccccccc}
				\toprule[1.5pt]
  \textbf{Wt Com} & \textbf{25\%}      & \textbf{50\%}      & \textbf{75\%}  & \textbf{Wt Com}       & \textbf{25\%}      & \textbf{50\%}      & \textbf{75\%}  
            \\\midrule[0.5pt]
\textcolor{blue}{1}0\textcolor{orange}{0}        & 41.89 & 44.03 & 42.42  &\textcolor{blue}{1}1\textcolor{orange}{0}           & 52.58 & 55.44 & 51.87    \\
\textcolor{blue}{0}1\textcolor{orange}{0} & 29.59 & 28.70 & 18.18  & \textcolor{blue}{1}1\textcolor{orange}{1}           & 21.75 & 20.14 & 15.15 \\
\textcolor{blue}{0}0\textcolor{orange}{1}      & 11.59 & 10.52 & 11.41  &\textcolor{blue}{2}1\textcolor{orange}{1}           & 42.96 & 45.10 & 44.39  \\
				\bottomrule[1.5pt]
		\end{tabular}}
	  \begin{tablenotes}
		\footnotesize
		\item[$*$] \textbf{Wt Com} means weight combination. The three columns of number represent the weights of the \textcolor{blue}{image}, motion vectors, and \textcolor{orange}{residual results}, respectively.
	\end{tablenotes}
	\end{threeparttable}
\end{table}

\textbf{Fusing multi-modal data aids accurate action recognition but requires a better trade-off of accuracy and efficiency.} 
There are two main approaches to fusing multi-modal data: fixed-weight fusion and adaptive fusion using learnable modules. We conduct thorough experiments on both methods to observe how fusing multi-modal data can achieve higher accuracy.
To reduce computational complexity, the SOTA in multi-modal action recognition \cite{guo2023lae} employs fixed weights for data fusion.
We perform inference on individual frames for frames located at the 25\%, 50\%, and 75\% positions (frame numbers approximately 19, 38, and 57), with varying weights to fuse the classification outcomes of three modalities. The results are shown in Table \ref{T.multi-modalDataResults}. 
Notably, while images perform best, integrating additional modalities significantly enhances accuracy.
Accuracy is 41.89\% with images alone, rising to 52.58\% with MVs and 45.28\% with Res.
However, when the results from all three modalities are fused with equal weights (1:1:1), the accuracy drops to 21.75\%. Increasing the weight of the image modality to 2 improves the accuracy back to 42.96\%. This indicates that there are too many noisein the MVs and residuals, and their inclusion in the fusion can degrade performance. 
Therefore, fixed weights cannot fully exploit the complementarity of multi-modal data, which limits the performance improvement of the fused model.

To fully tap into the potential of multi-modal data, some studies employ learnable fusion modules to automatically integrate features from various modalities.
Table \ref{T.learnableFusionModules} shows the evaluation of various representative multi-modal fusion modules regarding performance and accuracy. 
The first two methods are late-fusion methods, which fuse features at the backend of the network. Due to the small feature maps and low computational cost, these methods exhibit low running latency on edge devices. For instance, the CPU latency for Temporal Trilinear Pooling (TTP) \cite{huo2020lightweight} and Spatial Cascading and Pooling (SCP) \cite{zheng2023dynamic} is less than 1 ms. However, their accuracy still lags behind the fusion scheme with predefined weights (e.g., 1:0:0). 
This is because the small area of post-fusion features limits their ability to capture dominant features from each modality effectively.

In contrast, SIFP \cite{li2020slow} not only fuses final features but also exchanges intermediate features multiple times between modality branches. Consequently, among the learnable schemes, it achieves the highest accuracy gain at the cost of large latency, which is detrimental to the user experience in edge computing. 
 Learnable adaptive weights are challenging or require significant computational resources.
\textit{It inspires us that supplementing prior knowledge for each modality may combine the simplicity of fixed weight calculations with the flexibility of learnable post-fusion schemes, achieving a trade-off between accuracy and speed.}

\begin{table}[hbt]
	\setlength{\abovecaptionskip}{0pt}
\setlength{\belowcaptionskip}{1pt}
\renewcommand\arraystretch{1.2} 
	\caption{The accuracy(\%) and latency(ms) of learnable fusion modules.}
	\label{T.learnableFusionModules}
	\centering
	\begin{threeparttable}
		\setlength{\tabcolsep}{1.8mm}{
			\begin{tabular}{cccccc}
				\toprule[1.5pt] 
\multirow{2}{*}{Fusion Methods} & \multicolumn{3}{c}{\textbf{Accuracy}(\%)}                        & \multicolumn{2}{c}{\textbf{Latency}(ms)} \\
                         & 25\%       & 50\%    & 75\%                             & CPU          & GPU          \\
\midrule[0.5pt]
SIFP\cite{li2020slow}  & 56.34  & 58.72 & 57.07 &  81.40 & 261.89 \\ 
TTP\cite{huo2020lightweight} & 48.17 & 49.54 & 48.30    & 0.33  & 11.32  \\
SCP\cite{zheng2023dynamic} & 52.11& 53.76 &  49.02 &  0.64  & 5.44   \\
				\bottomrule[1.5pt]
		\end{tabular}}
	\end{threeparttable}
\end{table}

\section{\sysname Overview}
\label{sec:overview}

\begin{figure*}[hbtp]
  \centering
  \includegraphics[width=1.9\columnwidth]{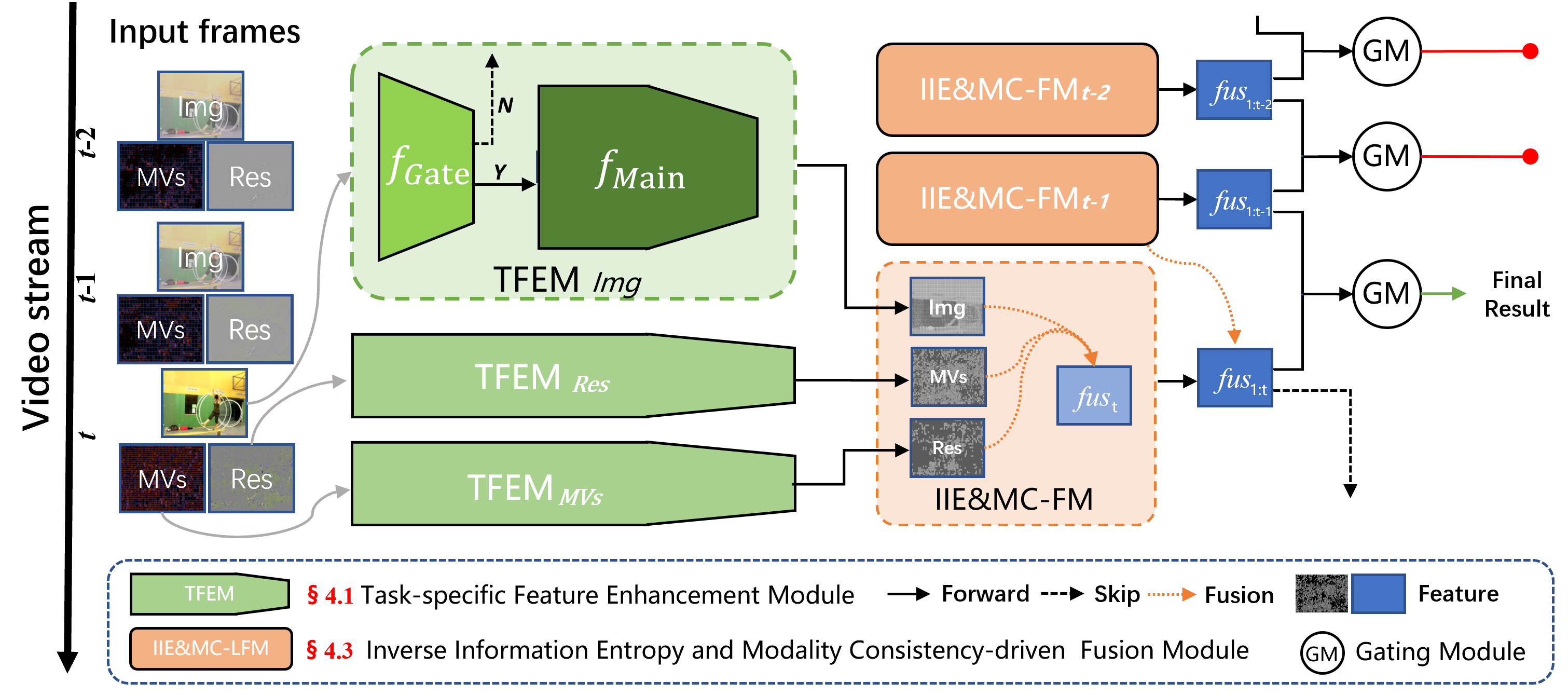}
  \caption{The workflow of \sysname.}
  \label{F.pipline}
\end{figure*}

\textbf{Design goal.}
The goal of \sysname is to achieve real-time online action recognition on edge devices, reducing user wait time. It includes two key aspects:
\begin{itemize}
    \item \textbf{Early exit:} 
    Without relying on the entire video information as a prior, it analyzes the video stream on edge devices and outputs results promptly after capturing sufficient action features.
    \item \textbf{Efficiency:} 
     \sysname should be edge-friendly to rapidly and efficiently model actions for each frame without incurring excessive performance overhead.    
\end{itemize}

\textbf{System Architecture.} 
Fig. \ref{F.pipline} shows the system architecture.
Each frame of the video stream is decoded to extract image and compressed-domain data. 
The extracted data is then fed into the corresponding branch of the Early Exit-oriented Task-specific Feature Enhancement Module (TFEM) (§\ref{Early Exit-oriented Task-specific Feature Enhancement Module}).
The TFEM branch consists of two stages, $f_{Gate}$ and $f_{Main}$. 
$f_{Main}$ activates to extract features and perform classification when $f_{Gate}$ deems the input worthy of inference.
To enable TFEM to learn the features at the beginning of the video, we design an iterative training($\S$.\ref{Iterative Training of fg and fm}) to alternately train $f_{Gate}$ and $f_{Main}$. 
$f_{Main}$ gradually adapts to the sampling field of view provided by $f_{Gate}$, and $f_{Gate}$ updates based on $f_{Main}$'s performance.

Subsequently, the features from different modalities are fed into the Inverse Information Entropy (IIE) and Modality Consistency (MC)-driven Fusion Module($\S$.\ref{IIE and MC-driven Fusion Module}). 
We design the IIE measure to assess the confidence of the TFEM for per modality and the MC weight evaluates inference reliability per frame.
We use the IIE and MC weights to guide the fusion of multi-modal features over time, yielding final features for classification and exiting prediction.

\section{Design}
\label{s.design}

In this section, we propose the key design of \sysname, aiming to achieve accurate online action recognition while maintaining competitive computational costs on edge devices. We first design TFEM to extract rich representations in both temporal and spatial dimensions. Subsequently,  we design a novel iterative training strategy to optimize \sysname, allowing the model to consciously capture features from real-time video streams, rather than requiring the complete video as a prior.
Finally, we design an Inverse Information Entropy(IIE) and Modality Consistency(MC)-driven Fusion Module for adaptive feature fusion in a learnable manner.

\subsection{Early Exit-oriented Task-specific Feature Enhancement Module (TFEM)}
\label{Early Exit-oriented Task-specific Feature Enhancement Module}

The task-specific feature enhancement module (TFEM) consists of sequential stages: $f_{Gate}$ and $f_{Main}$. $f_{Gate}$ performs initial filtering of the input data. 
When $f_{Gate}$ verifies the value of the current frame to determine the action category, $f_{Main}$ is activated.
$f_{Main}$ conducts a detailed analysis of the input frames and generates the final classification results. 
Specifically, we design two plug-and-play lightweight modules for $f_{Gate}$ and $f_{Main}$ to enhance their temporal and spatial feature extraction capabilities in online scenarios, respectively. 
These modules can be inserted into any backbone model for action recognition.

\textbf{Temporal Layering Shift Module (TLSM).}
We design TLSM to share information between consecutive frames in a video stream thereby enhance the temporal feature representation of $f_{Gate}$. It swaps channels between past frames, from distant frames to recent ones, as described by Equ.\ref{E.TLSM}.
\begin{equation}
\label{E.TLSM}
\mathbf{x}_{t}[:,:,:c]=Cat_{j=1}^n \mathbf{x}_{t-j}[:,:,: \frac{c}{n} ]
\end{equation}
where $\mathbf{x}_{t}$ represents the features of the $t$-th frame, $n$ is the total number of preceding frames considered, $c$ is the number of channels to be shifted, and $Cat$ denotes the concatenation operation.

\begin{figure}[htb]
\centering
\captionsetup[subfloat]{labelfont=small,textfont=normalsize,farskip=0pt}
\subfloat[Accuracy]{
\includegraphics[width=0.49\columnwidth]{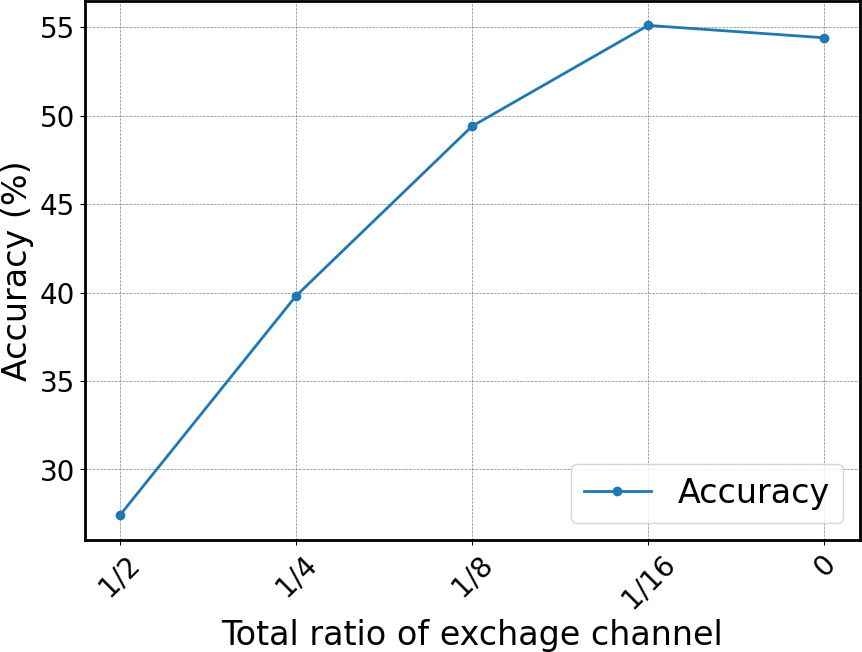}
}
\subfloat[Lantency]{\includegraphics[width=0.49\columnwidth]{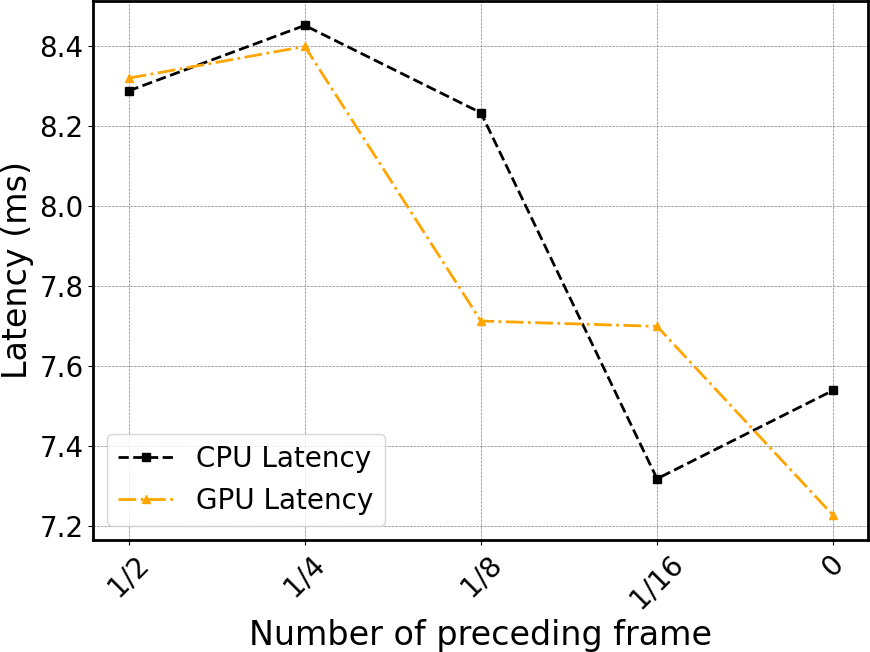}}%
\caption{Exchange ratio of channels \textit{vs.} accuracy and lantency. Number of preceding frame = 4.
} 
\label{F.exchangechannle}
\end{figure}

\begin{figure}[htb]
\centering
\captionsetup[subfloat]{labelfont=small,textfont=normalsize,farskip=0pt}
\subfloat[Accuracy]{
\includegraphics[width=0.49\columnwidth]{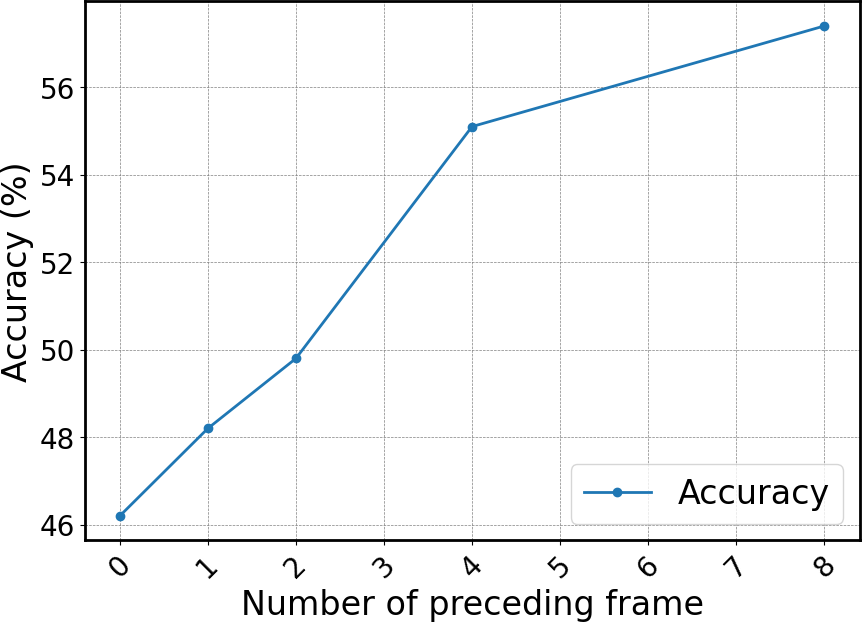}
}
\subfloat[Lantency]{\includegraphics[width=0.49\columnwidth]{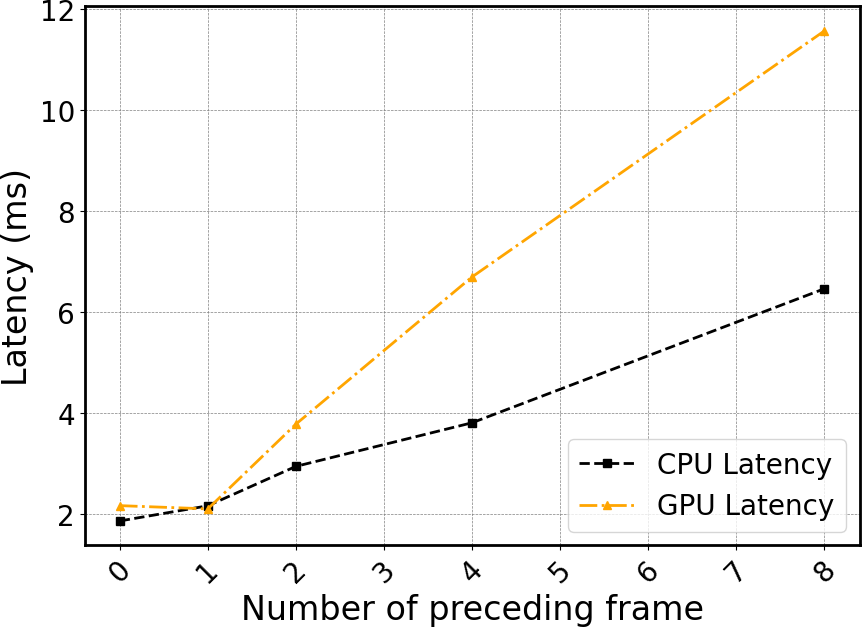}}%
\caption{Number of shared frame \textit{vs.} accuracy and lantency. Ratio of channels = 1/16.
} 
\label{F.shareframe}
\end{figure}

We evaluate the impact of varying values of $c$ and $n$ on HDMB51 with ShuffleNet V2. As shown in Fig.\ref{F.exchangechannle}a, shifting a small number of channels yields a notable improvement in accuracy compared to not moving channels. However, accuracy diminishes as the proportion of channel movement increases. This decline is attributed to the interference with the feature learning of the current frame due to the introduction of too many channels from other frames. 
In Fig.\ref{F.exchangechannle}b, the average operation latency is basically unaffected, and the exchange of different numbers of channels only increases the latency by approximately 1 ms.
Subsequently, we fixed the exchange channel ratio at 1/16 and assessed the performance impact of utilizing different ranges of previous frames. As shown in Fig.\ref{F.shareframe}, when $n = 1$, features are only shared from the immediately preceding frame. Since adjacent frames are highly similar, this does not substantially enhance performance. 
We find that expanding the feature-sharing range, especially between more distant frames, notably enhances the model's temporal modeling capabilities.
Increasing the number of shared frames from 4 to 8 resulted in only a 4.17\% improvement in accuracy, at the cost of a 24.3\% and 19.94\% increase in CPU and GPU latency, respectively. Consequently, we set the number of shared frames $n$ to 4.

\textbf{MBs-guided Spatial Enhancement Module (MSEM).}
To boost $f_{Main}$'s spatial feature extraction, we designed the MBs-guided Spatial Enhancement Module (MSEM) to guide $f_{Main}$ to focus spatially semantic-rich areas.
It uses the intermediate results from the video compression process on edge devices, i.e., MBs, as an auxiliary input. 
The encoder delineates smaller MBs in areas of complex scene content and motion variation, which can indicate spatio-temporal features of motion to some extent. We employ a mapping scheme to quantify the area of MBs within a frame. The value at position $(i,j)$ of the MBs map $\mathbf{x}^{MB}$ is determined according to Equ.\ref{E.mbvalue}:
\begin{equation}
\label{E.mbvalue}
    \mathbf{x}^{MB}(i,j)=1-norm(\log_{2}{S_{(i,j)}})
\end{equation}
where $S_{(i,j)}$ is the area size of the MBs containing the pixel at position $(i,j)$. The function $\text{norm}(\cdot)$ denotes min-max normalization. 
Under the H.265 standard, MBs areas range from $2 \times 2$ to $64 \times 64$.
We apply a logarithmic function to reduce the size disparity and use $1 - \text{norm}(\cdot)$ to prioritize smaller macroblocks with higher values. 
We formulate the MSEM as Equ.\ref{E.sae_mb}.
\begin{equation}
\label{E.sae_mb}
\mathbf{x}^{\prime}_{i}= \sigma \left( Conv ([mean(\mathbf{x}_t^{MB});max(\mathbf{x}_t^{MB})]) \right)  \odot \mathbf{x}_{t}
\end{equation}
where $ \mathbf{x}_t $ and $ \mathbf{x}_t^{MB}$ represent the features and MBs map of the $t$-th frame, respectively. We linearly scale the MBs map to match the size of $ \mathbf{x}_t $. The function $ \sigma $ denotes the Sigmoid function, and $ Conv(\cdot) $, $ mean(\cdot) $, and $ max(\cdot) $ represent convolution, average pooling, and max pooling operations, respectively. The symbol $ [\cdot; \cdot] $ indicates concatenation along the channel dimension, and $ \odot $ denotes element-wise multiplication.

\subsection{Iterative Training of $f_{Gate}$ and $f_{Main}$}
\label{Iterative Training of fg and fm}
To effectively optimize \sysname for learning the key features from the initial frames of a complete video, we propose a novel progressive training strategy. This strategy aims to enable the model to rapidly achieve accurate action recognition with high confidence during real-time inference. Specifically, we first use an iterative training process to refine TFEM weights. 

\textbf{Iterative TFEM Weight Optimization.}
\label{Iterative TFEM Weight Optimization.}
To ensure that $f_{Gate}$ and $f_{Main}$ work well for online video streams, we design an iterative TFEM weight optimization process to bridge the gap of frame sampling between training and online inference.

To initialize the weights of TFEM, we pre-train $f_{Main}$ by segmenting a video and randomly sampling  a frame from each segment as described  in \cite{wang2016temporal}. 
To enable the information transfer between $f_{Gate}$ and $f_{Main}$, we define a hard label $\tilde{y}$. 
$\tilde{y}=1$ if $f_{Main}$'s output matches the ground truth at the frame level; otherwise, $\tilde{y}=0$. We then optimize $f_{Gate}$ using the hard label. 

\begin{figure}[htb]
	\centering \includegraphics[width=1\columnwidth]{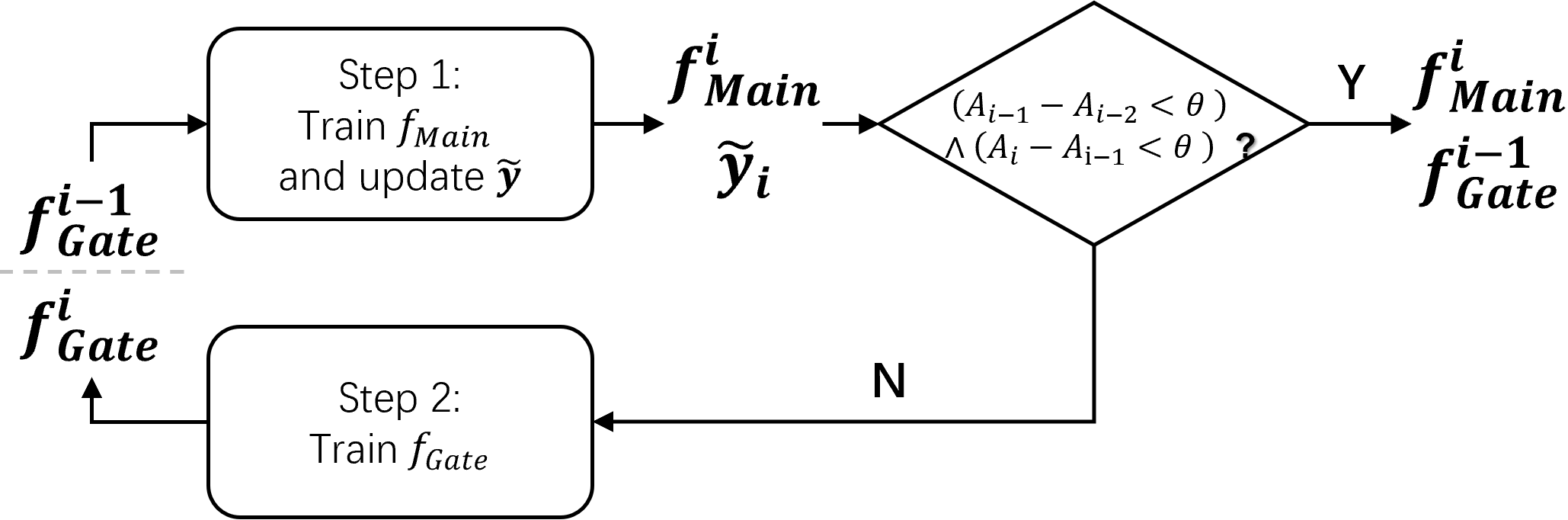}
	\caption{The iterative training process of TFEM in the $i$-th iteration. 
 }
	\label{F.cascadednetwork}
\end{figure}

In the following, we elaborate the iterative training process of $f_{Main}$ and $f_{Gate}$.

\begin{enumerate}
    \item Train $f_{Main}^{i-1}$ with frozen $f_{Gate}^{i-1}$. 
    We randomly sample frames from the list provided by $f_{Gate}^{i-1}$, specifically those with $f_{Gate}^{i-1}$ output equal to 1.
    After every three training epochs, we perform a test. If the increase in accuracy between two consecutive tests is less than $\theta$, we update the hard label $\tilde{y}_{i-1}$ to $\tilde{y}_{i}$. If the accuracy satisfies the stopping criterion, i.e., Eqn. \ref{E.stopiteration}, output the $f_{Main}^i$ and $f_{Gate}^{i-1}$.  Otherwise, go to \textit{Step 2}.

    \begin{equation}
    \label{E.stopiteration}
        (A_{i-1}-A_{i-2}<\theta) \land (A _{i}-A_{i-1} <\theta)
    \end{equation}
    where, $A_{i}$ is the accuracy of $f_{Main}^i$. We empirically set $\theta = 10^{-2}$,

    \item Use the updated hard label $\tilde{y}_i$ to train $f_{Gate}^{i-1}$. 
    We apply the stopping criterion from \textit{Step 1}.
    We perform a test after every three training epochs. 
    If the increase in accuracy in adjacent two tests is less than $\theta$, we output the $f_{Gate}^i$ and go to \textit{Step 1}.
\end{enumerate}

\textbf{Temporally Decaying Prioritization(TDP).}
We employ TDP, as defined by Equ.\ref{E.TDP}, to prioritize earlier frames and enhance TFEM's focus on initial segments.
According to Equ.\ref{E.TDP}, the first $\tau$ frames receive half of the total priority. Subsequently, each following frame is allocated half of the remaining priority. This process continues until the last frame, which receives the entirety of the remaining priority. This setting guarantees that the priorities of all sampled frames sum to 1, thereby obviating the need for additional normalization. 
\begin{equation}
\label{E.TDP}
TDP_t=\begin{cases}
1/2^\tau \quad &t\le\tau \\
1/2^{t-\tau+1}\quad &t>\tau \text{ \& }t<v_{len} \\
1/2^{t-\tau+2}\quad &t=v_{len}
\end{cases}
\end{equation}
where $t$ denotes the index of a frame, $v_{len}$ is the total number of sampled frames. Based on our experiments, we empirically set $\tau = 2$.

TDP is applied in iterative training in two key ways:
\begin{itemize}
    \item 
    During training of $f_{Main}$ and $f_{Gate}$, we adjust the priorities of the extracted features from $ v_{\text{len}}$ frames and set their priorities according to Equ.\ref{E.TDP}.
    \item 
    When calculating accuracy, we process results from each frame using Equ.\ref{E.TDP}, summing them to obtain the final accuracy, where $v_{\text{len}}$ equals the video length.
\end{itemize}

\subsection{Inverse Information Entropy(IIE) and Modality Consistency(MC)-driven Fusion Module}
\label{IIE and MC-driven Fusion Module}
To enhance online inference accuracy, we first design the IIE measure to assess TFEM confidence per modality.
Then, we employ a learnable module to determine modality fusion weights, reflecting each branch's contribution and guiding multi-modal feature fusion.
Building on this, we utilize MC to assess the reliability of inference results per frame and guide temporal feature fusion.
We extract the fused features from the first to the current frame for classification and stop prediction.

\textbf{IIE Measure-driven Modal Weight.}
To quantify the confidence of TFEM's output, we propose the IIE measure. 
Intuitively, a clear peak in TFEM's output distribution $P=[\rho_1,\rho_2,\dots,\rho_N]$ indicates high model confidence in classifying the input sample.
Conversely, if the distribution is flat or spread out, it suggests that the model is uncertain and indecisive among multiple classes, reflecting higher uncertainty in the classification result. Driven by this intuition, we design the IIE as in Equ.\ref{E.IIE}:

\begin{equation}
\label{E.IIE}
IIE = 1 - \left( -\sum_{i=1}^{N} \rho _i \log_2 \rho_i \right) / \log_2(N)
\end{equation}
where, $N$ is the total number of action categories. $\sum_{i=1}^{N}\rho_i\log\rho_i$ denotes the information entropy. By dividing the information entropy by $\log_2(N)$, we normalize the information entropy to a value between 0 and 1. Finally, we obtain the IIE measure by subtracting the normalized information entropy from 1. Higher IIE values indicate greater confidence of TFEM. 

Specifically, we use a learnable modality fusion module to refine the weights of the multi-modal TFEM branches. 
First, each modality's $t$-th frame is input into its branch network to extract feature vector $\mathbf{x}_t$ and classification probability vector $P$.
Next, we employ the IIE measure, to integrate features from different modalities. 
The module generates weighted features $\mathbf{w}$ for each modality via a weight fusion process, as described by Equ.\ref{E.WFM}. 

\begin{equation}
\label{E.WFM}
   \mathbf{w} = \sigma \left( Conv (\mathbf{x}_{t}\times IIE) \right) 
\end{equation}
where $\mathbf{w}$ is the learnable weight map for some modality. 
Then, we get the weighted features by $\mathbf{x}^{'}_{t}= \mathbf{w} \odot \mathbf{x}_{t}$, $\odot$ denotes element-wise multiplication.

Finally, we concatenate the processed features from each modality and pass them through a $1 \times 1$ convolution to produce the fused feature map $\mathbf{x}^{fus}_t$. We then feed this fused feature into a classifier to yield the final result.
Compared to directly using IIE as fusion weights, learnable modal weights allow local fine-tuning of feature weights $\mathbf{w}$.
This approach results in weights that better match  the characteristics of each modality.


\textbf{MC-driven Frame Weight.}
We design the Modality Consistency-driven (MC) frame weights to quantify this intuition. First, we calculate the average value  $\bar{\omega}^M$  of the learned weight map $\mathbf{w}^M$ of modality $M$. Then, we identify the maximum weight within each modality. Among all $m$ modalities, the one with the highest weight is recognized as the salient modality $\bar{\omega}^{salient}$.
Based on this, we define the MC weights $w_{mc}$ for the frame as Eqn.\ref{E.mcweight}

\begin{equation}
\label{E.mcweight}
w_{mc}=\bar{\omega}^{salient}+Sign^{m-1}_{j=1}(\bar{\omega}^j)
\end{equation}
where $m$ represents the total number of modalities used, and $Sign$ is a function we define. We compare the inference results of non-significant weighted modality branches with those of the branch associated with $\bar{\omega}^{salient}$. 
The output is 1 if inference results match; otherwise, it is -1.
Here, we determine the sign of each branch's weights based on inference consistency, obtaining the frame weight $w_{mc}$.
Subsequently, we compute the weighted average of the features from the first frame to the current frame based on their respective MC weights, thereby obtaining the weighted fused features up to the current frame $x^{fux}_{1:t}$ as Eqn.\ref{E.fus2}.

\begin{equation}
\label{E.fus2}
\mathbf{x}^{fux}_{1:t}=\frac{1}{t}\sum^t_{i=1}{w_{mc_i}\times \mathbf{x}^{fux}_{i}}
\end{equation}

Note that we set the modality's $\bar{\omega} ^{'} = 0$ if $f_{Main}$ in a modality is not active in the current frame. Unlike common fusion methods such as weighted averaging \cite{wang2016temporal,wu2018compressed}, this approach considers modality consistency
and enhancing accuracy in multi-modal decision-making.

\textbf{Gating module.}
Finally, we input the current frame's fused feature $x^{fux}_{1:t}$ and the previous frame's fused feature $\mathbf{x}^{fux}_{1:t-1}$ into the gating module, as defined in Equ.\ref{E.GM} to predict the \sysname stopping confidence.
\begin{equation}
\label{E.GM}
B =\sigma (\Pi ([FC(\mathbf{x}^{fus}_{1:t-1}):FC(\mathbf{x}^{fus}_{1:t})] ))
\end{equation}
where $FC$ denotes a fully connected layer, $\Pi$ denotes linear projection.
When $ t = 1$, we copy the fused features $\mathbf{x}^{fus}_{1}$ to facilitate the computation due to the absence of features from the previous frame.
The result is passed through the sigmoid function $\sigma$ to get the gating result $B \in {0,1}$. We stop processing if $B = 1$; otherwise, we continue with subsequent frames until the video's last frame.

We set the label $y^g_i$ to 1 to encourage early final prediction if the following two conditions are met; otherwise, we set it to 0.
First, the recognition result $\mathbf{x}_i^{fus}$ must match the ground truth. Second, $y^g_{i} = 1$ can occur no more than 5 times.
\section{Implementation}
\label{s.implementation}

We implement \sysname in Java \cite{arnold2005java} on Android edge devices. 
Our implementation utilizes Android's MediaCodec interface \cite{mediacodec} for H.265 video encoding and FFmpeg \cite{FFmpegmain} for video decoding and processing.
This approach achieves efficient hardware acceleration and multithreading, ensuring fast processing times without disrupting FFmpeg's decoding processes or its optimization and acceleration techniques.
\sysname is trained using PyTorch on a server equipped with an GeForce RTX 3090 GPU card and 180 GB of memory. 

\textbf{Backbones of TFEM.} We utilize ShuffleNet V2\cite{ma2018shufflenet} as the backbone for $f_{Gate}$. 
The synergy of group convolutions and TLSM's temporal extension boosts model performance on video and sequential image data.
We utilize the ResNet-50\cite{he2016deep} as the backbone for $f_{Main}$. 
We add TLSM to the first convolutional layer which stride is 1 in the second branch in ShuffleNet V2, and add MSEM after the first residual blocks in ResNet.

\textbf{Deployment on edge devices.} All neural network modules are deployed using the MNN framework\cite{mnn}. Models are exported from PyTorch\cite{Pytorch} and converted into supported MNN format (.mnn), then load through the Android Java JNI interface for efficient CPU and GPU inference. All models were deployed using FP16 precision. For each decoded frame, we first use the accuracy selector for inference; if the output is 1, it is then fed into the classification model. Different modalities of the model utilize CPU and GPU inference based on availability to optimize inference speed.

\textbf{Data format setting.} We re-encode the videos in our dataset on Snapdragon 865 devices into H.265 format. 
The Group of Pictures (GOP) size is set to 12, with a medium bitrate and no B-frames.
We modify FFmpeg to decode images and compressed domain information from the videos. We interpolate the MVs to match the size of the images and store them as two-dimensional arrays. MVs and Res of a single frame only encode the differences relative to the adjacent frames.
Variations in these changes lead to significant differences in adjacent MVs and residuals.
To enhance frame change impact, we accumulate compressed domain data within each GOP, implementing Coviar\cite{wu2018compressed}.

\textbf{Training Setting.}
The pre-training parameters for TFEM's backbones in both stages are: 50 epochs, initial learning rate 0.01 (decayed by 0.1 at epochs 20 and 40), weight decay 1e-4, batch size 64, and dropout rate 0.5. 
TFEM's backbones are pre-trained on kinetics-400\cite{carreira2017quo}  
For subsequent training, unless specified, the learning rate is 0.01, and other parameters are unchanged.
We optimize the parameters of the backbone and classifier for both $f_{Gate}$ and the gating module using binary cross-entropy loss. 
The parameters for $f_{Main}$ and fusion modules are optimized using cross-entropy loss.

\section{Evaluation}

\label{s.evaluation}

In this section, we evaluate the effectiveness of \sysname in the following steps. 
We first present the evaluation setup, then analyze the overall performance and provide a detailed analysis of each key component, including latency, accuracy, and energy consumption. We also compare \sysname with other state-of-the-art methods.

\subsection{Experiment setup}
\textbf{Hardware Platforms.} We conducted testing on four edge devices proposed in recent years. 
These devices encompass a range of computational capabilities, as shown in Table.\ref{T.Hardware}. 
To ensure consistent System on Chip(SoC) performance, devices are fully charged to 100\% and disconnected from the power supply during speed tests. Devices are rebooted before each test to clear caches and background processes. 
The target model or algorithm is warmed up with 10 runs then executed 100 times for data collection. The reported results is the average of these 100 runs after the warmed up.

\begin{table}[hbt]
	\setlength{\abovecaptionskip}{0pt}
\setlength{\belowcaptionskip}{1pt}
\renewcommand\arraystretch{1.2} 
	\caption{Hardware configurations of the edge devices used in the experiments.}
	\label{T.Hardware}
	\centering
	\begin{threeparttable}
		\setlength{\tabcolsep}{1.8mm}{
			\begin{tabular}{cccc}
				\toprule[1.5pt]
				\textbf{Device\tnote{$*$}} & \textbf{SoC} & \textbf{CPU(big core)} & \textbf{GPU} \\  \midrule[0.5pt]
XM14P  & Snapdragon   8g3\cite{snapdragon_8g3}  & Cortex X4      & Adreno 750         \\
RM12T & Snapdragon   7+g2\cite{snapdragon_7g2} & Cortex X2 & Adreno 725         \\
RMk30  & Snapdragon   865\cite{snapdragon_865}  & Cortex A77      & Adreno 650         \\
HW40P & Kirin 9000\cite{kirin_9000}          & Cortex A77\&A55       & Mali G78 MP24         \\
				\bottomrule[1.5pt]
		\end{tabular}}
	  \begin{tablenotes}
		\footnotesize
		\item[$*$]XM14P, RM12T, RMk30, HW40P are abbreviations of Xiaomi 14 Pro, Redmi Note 12 Turbo, Redmi K30, Huawei Mate 40 Pro respectively.
	   \end{tablenotes}
	\end{threeparttable}
\end{table}

\textbf{Datasets.} 
We evaluate the performance of \sysname using two public action recognition datasets: UCF-101\cite{soomro2012datasetucf101} comprising 13,320 video clips across 101 action categories, and HMDB-51\cite{kuehne2011hmdb} containing 6,766 clips divided into 51 action categories.
These datasets cover a broad spectrum of actions, filmed in diverse settings (indoor and outdoor) with varied camera movements (static, tracking, handheld).
We retain the original video resolution and re-encode them into H.265 format, setting GOP to 12, FPS to 30, using one reference frame, and excluding B-frames.

\textbf{Baselines.}
We compare \sysname with representative baselines on two datasets. These include VideoMAE-2 \cite{wang2023videomae}, TSN\cite{wang2018temporal} , FrameExit\cite{ghodrati2021frameexit} and TLEE \cite{wang2023tlee}, they inference only images. Among them, FrameExit and TLEE do not infer all sampled frames, because them perform early exits to save computing resources.
LAE-Net\cite{guo2023lae}, Coviar\cite{wu2018compressed}, TTP\cite{huo2020lightweight}, CoviFocus\cite{zheng2023dynamic} and SIFP\cite{li2020slow} all use multi-modal for inference. 
Except that CoviFocus uses two modalities of Imgs and MVs, the other methods use all three modalities. 
All the above methods are designed for offline recognition.
The sampling frame of VideoMAE-2 and LAE-Net is 16, other is 10.
The backbones of VideoMAE, LAE-Net, TTP and CoviFocus are ViT-Base\cite{dosovitskiy2020image}, Resnet-152\cite{he2016deep}, Mobilenet-V2\cite{sandler2018mobilenetv2} and Mobilenet-V2 + Resnet-50\cite{he2016deep} respectively. The backbone of other methods including \textit{ours} is Resnet-50.

To better adapt to real-world scenarios, we modify FrameExit and TLEE to perform frame-by-frame inference instead of relying on globally sampled frames. During training, we randomly sample 10 frames from any video segment. For inference, we conduct frame-by-frame analysis, aligning with the approach of \sysname.

\textbf{Evaluation Metrics.} 
We evaluate these methods runtime using latency, timing from the start of video playback until the result is generated. 
We report Top-1 accuracy for performance evaluation.\cite{wang2018temporal}
These metrics range from 0-1, with higher values indicating better segmentation performance.
During testing, to better reflect the model's ability to capture features from the beginning of the video, we utilized TDP, which is calculated according to Equ.\ref{E.TDP}.
We report the energy consumption of these methods from framework startup to result generation, excluding the energy used for video playback, to fairly comparison the framework’s energy efficiency.

\subsection{Overall Performance}

\begin{figure}
\centering
\captionsetup[subfloat]{labelfont=small,textfont=normalsize,farskip=0pt}
\subfloat[UCF-101]{
    \begin{minipage}{1\columnwidth}
    \includegraphics[width=1\columnwidth]{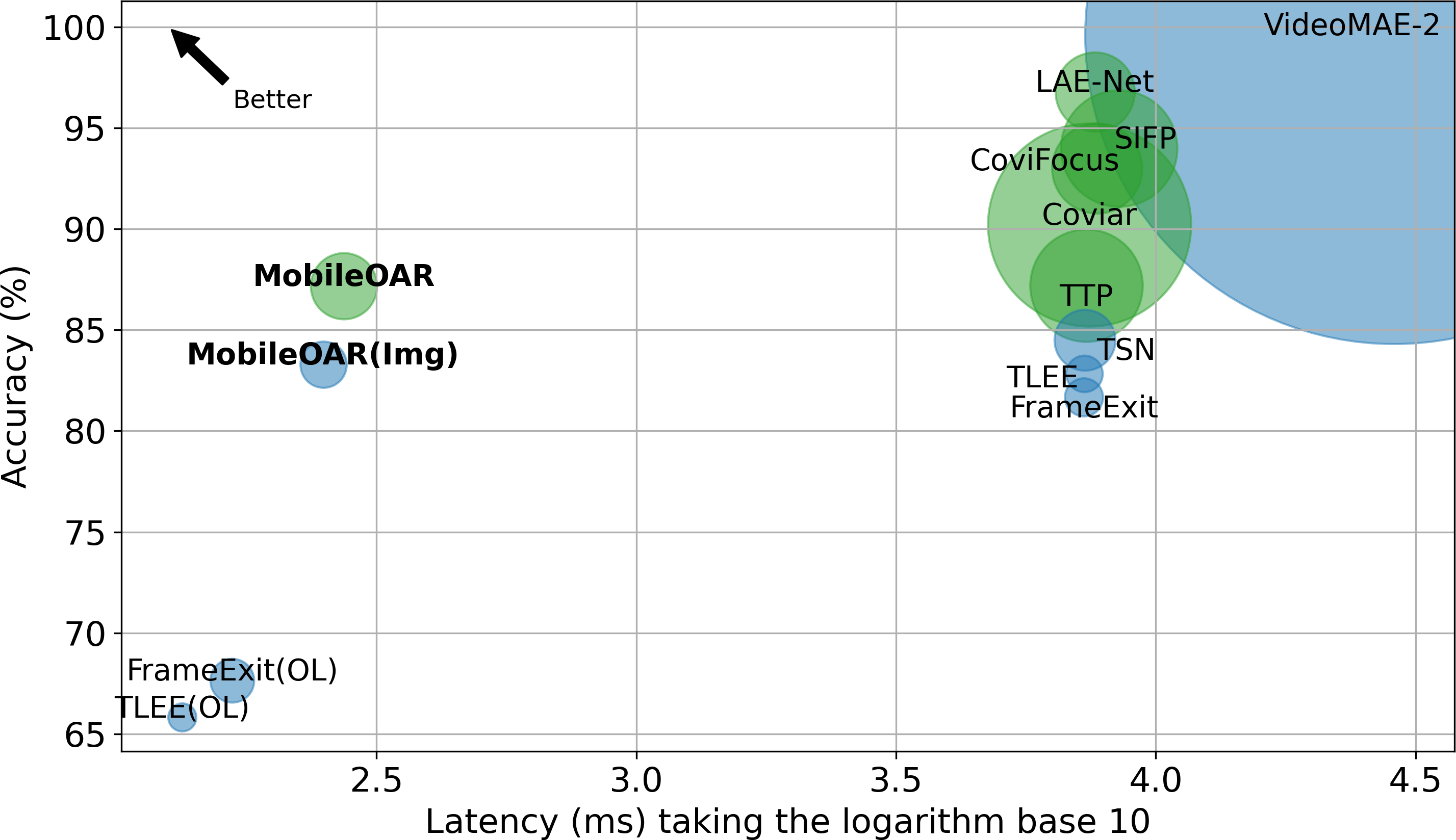} 
    \end{minipage}
}
	
\subfloat[HDMB51]{
    \begin{minipage}{1\columnwidth}
    \includegraphics[width=1\columnwidth]{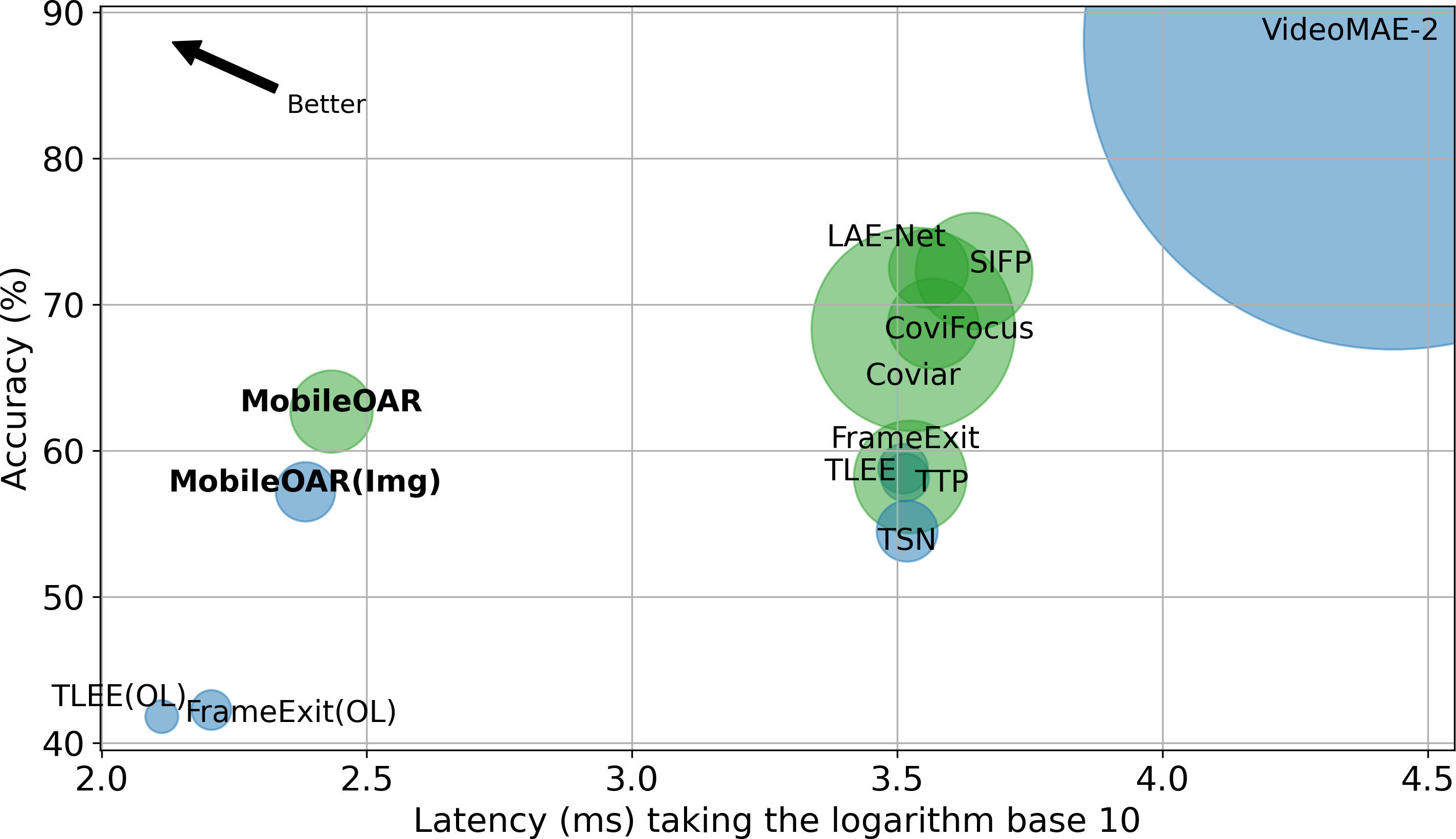} 
    \end{minipage}
} 
 \caption{
Latency \textit{vs.} accuracy on UCF-101 and HMDB-51, with bubble size representing average FLOPs per video. Blue bubbles representing image-only methods; green bubbles representing multi-modal data.
}
 \label{F.latency_acc}
\end{figure}

We benchmark \sysname against a range of action recognition methods, evaluating them based on accuracy and latency. 
As shown in Fig.\ref{F.latency_acc}, the $y$-axis represents inference accuracy, the $x$-axis represents inference latency, and bubble size indicates model FLOPs. Blue bubbles denote methods using only images, and the green ones represent using multi-modal data. 
In general, online methods significantly outperform offline methods in terms of latency by terminating and delivering results earlier. 
This advantage arises because offline schemes must wait for the video playback to complete before initiating processing, which has a negative impact on the user experience. 

In Fig.\ref{F.latency_acc}, our method occupies the upper left corner, indicating low latency while maintaining satisfactory accuracy. This makes it suitable for edge applications such as environmental anomaly detection and motion recognition \cite{zhang2019towards, wang2023tlee}. Although our method shows a gap in accuracy compared to VideoMAE-V2, it demonstrates a significant latency advantage, enhancing its competitiveness in practical scenarios. Specifically, compared to all offline methods, \sysname reduces latency by an average of 98.25\% and 96.23\%, with only a 2.20\% and 2.06\% decrease in accuracy, respectively. 

Our accuracy improves compared to some offline methods, i.e., TSN and TTP. In particular, compared to TSN, our accuracy increases by 14.98\% and 14.98\%, respectively, and compared to TTP, the accuracy increases by 0.02\% and 7.70\%, respectively. 
The improvement in accuracy stems from two fold. First, our MSEM use additional spatial information, enhancing the feature extraction capability of $f_{Main}$. Additionally, the IIE and MC-driven Fusion Module integrate valuable information from non-image modalities, further boosting accuracy. 
Second, our iterative training scheme that enhancing the ability of TFEM to extract the frame's features at the beginning of the video. 

Compared to online methods, both the modified FrameExit and TLEE exhibit a significant decrease in accuracy due to premature exits. \sysname improves average accuracy by 16.41\% and 36.53\% compared to these methods. This is because dense sampling can lead to information overload and confusion in their gating modules, resulting in premature convergence and early prediction failure. In contrast, our method balances accuracy and speed, highlighting the effectiveness of our gated stop condition.


\begin{table}[hbt]
	\setlength{\abovecaptionskip}{0pt}
\setlength{\belowcaptionskip}{1pt}
\renewcommand\arraystretch{1.2} 
	\caption{Energy consumption comparison on different devices (W·s).}
	\label{T.energy}
	\centering
	\begin{threeparttable}
		\setlength{\tabcolsep}{1.8mm}{
			\begin{tabular}{clcccc}
				\toprule[1.5pt]
\textbf{Notes}                        & \textbf{Method}         & \textbf{XM14P}    & \textbf{RM12T}    & \textbf{RMk30}    & \textbf{HW40P}    \\\midrule[0.5pt]
\multirow{2}{*}{\shortstack{Image\\only}}  & VideoMAE-2\cite{wang2023videomae}     & 219.23 & 238.53 & 208.59 & 216.09 \\
                             & TSN\cite{wang2018temporal}            & 1.27   & 1.05   & 0.69   & 1.11   \\
\hdashline
\multirow{5}{*}{\shortstack{Multi\\modal}} & LAE-Net\cite{guo2023lae}        & 4.71   & 3.84   & 2.40   & 4.64   \\

                             & Coviar\cite{wu2018compressed} & 8.39   & 3.36   & 0.85   & 9.45   \\
                             & TTP\cite{huo2020lightweight}            & 1.43   & 1.20   & 0.86   & 2.56   \\
                             & CoviFocus\cite{zheng2023dynamic}      & 4.76   & 4.20   & 2.70   & 5.87   \\
                             & SIFP\cite{li2020slow}           & 8.95   & 8.41   & 5.84   & 15.70  \\
\hdashline
\multirow{2}{*}{\shortstack{Early\\exit}}  & FrameExit\cite{ghodrati2021frameexit}      & 1.00   & 0.82   & 0.51   & 0.83   \\
                             & TLEE\cite{wang2023tlee}           & 0.91   & 0.68   & 0.43   & 0.72   \\
\hdashline
\multirow{4}{*}{Online}      & FrameExit(OL)  & 0.79   & 0.24   & 0.37   & 0.55   \\
                             & TLEE(OL)       & 1.23   & 0.21   & 0.30   & 0.43   \\
                             & \textbf{EdgeOAR}(Img) & 1.26   & 0.46   & 0.56   & 0.90   \\
                             & \textbf{EdgeOAR}      & 1.81   & 1.63   & 0.97   & 1.89  \\
				\bottomrule[1.5pt]
		\end{tabular}
        }
	\end{threeparttable}
\end{table}


Energy consumption of the solutions is primarily influenced by the number of sampled frames and the complexity of the backbone.
As shown in Figure \ref{T.energy}, VideoMAE-V2, which uses ViT-Base (263.93 GFLOPs) as the backbone and a dense sampling rate of 16 frames, achieves the highest accuracy but also consumes the most energy. In comparison, \sysname reduces energy consumption by an average of 99.29\%, due to the use of a much smaller backbone. 
Compared to LAE-Net, which achieves the best accuracy using compressed domain information, \sysname reduces energy consumption by an average of 59.45\%. Although both methods use ResNet-50 as the backbone, LAE-Net samples 16 frames, while \sysname employs a more flexible sampling strategy.

Compared to other multi-modal methods, \sysname reduces energy consumption by an average of 43.33\%. For example, compared to Coviar and CoviFocus, \sysname reduces energy consumption by 48.89\% and 63.73\%, respectively, because it uses a smaller backbone. Compared to SIFP, \sysname reduces energy consumption by 82.92\%, as SIFP performs multiple feature interactions and fusions between modality branches, increasing the computational load. In comparison to TTP, \sysname increases energy consumption by an average of 12.33\%, primarily due to TTP's use of a smaller backbone, MobileNet-v2.

Compared to TSN, FrameExit, and TLEE, \sysname increases energy consumption by an average of 52.57\%, 98.93\%, and 131\%, respectively. This is because these methods only use images to inference, and the latter two methods additional employ early exits, further reducing inference time. Compared to these methods and \sysname (Img), \sysname reduces energy consumption by an average of 23.57\% and 0.4\%, respectively, with only a 14.89\% increase relative to TLEE. This indicates that the increased energy consumption in \sysname is mainly due to the need to infer more modalities.

Compared to the modified online methods, FrameExit(OL) and TLEE(OL), \sysname(Img) increases energy consumption by an average of 65.63\% and 77.53\%, respectively. This is because the gated stop condition in \sysname allows for viewing more frames to achieve better accuracy.

\subsection{Performance Breakdown}

Next, we meticulously assess the performance of key components within the \sysname framework. All results presented in this section were obtained through testing on the UCF-101 dataset.

\begin{figure}[htb]
\centering
\captionsetup[subfloat]{labelfont=small,textfont=normalsize,farskip=0pt}
\subfloat[Played frames]{
\includegraphics[width=0.50\columnwidth]{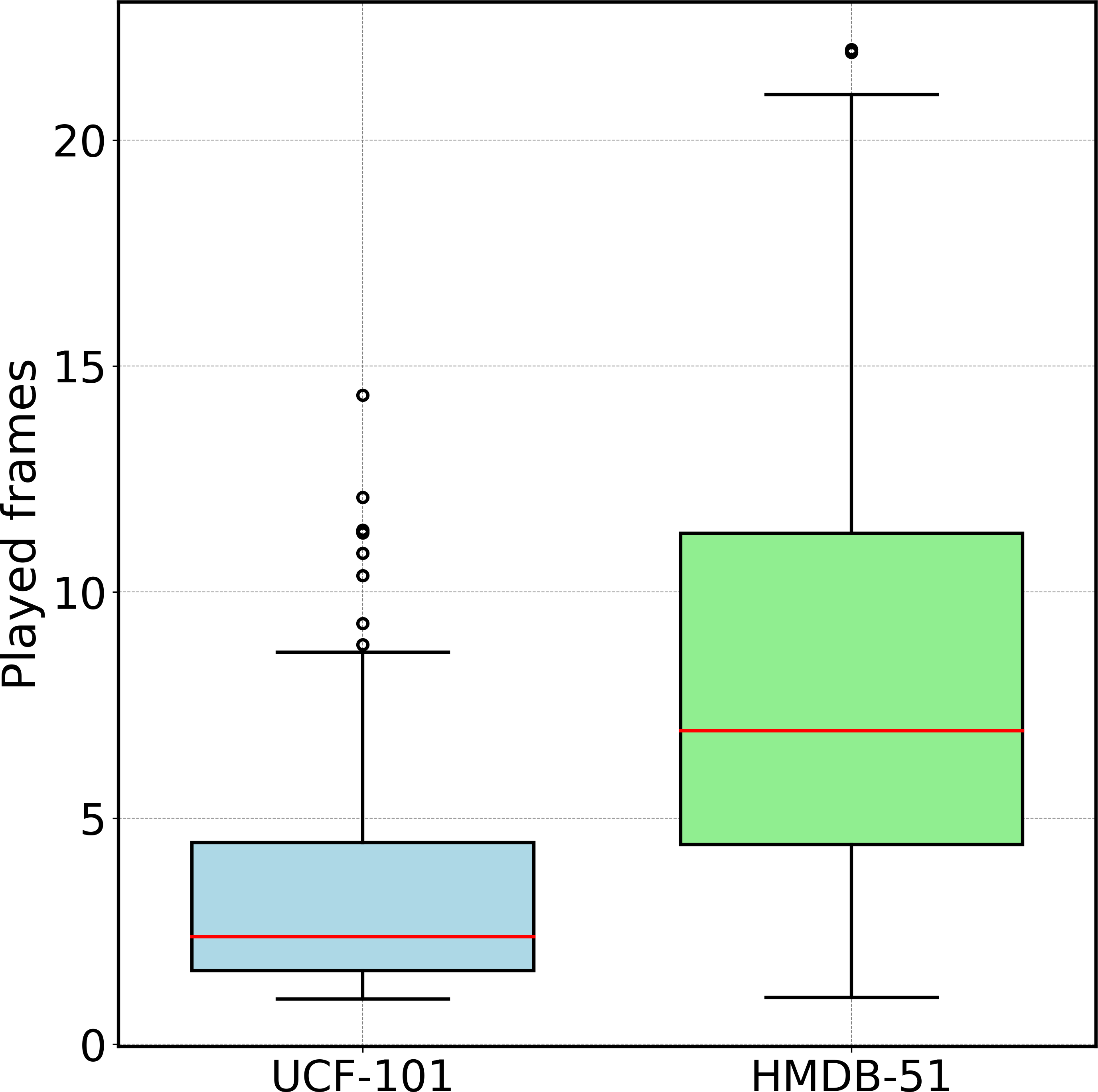}
}
\subfloat[$f_{Main}$ / $f_{Gate}$]{\includegraphics[width=0.50\columnwidth]{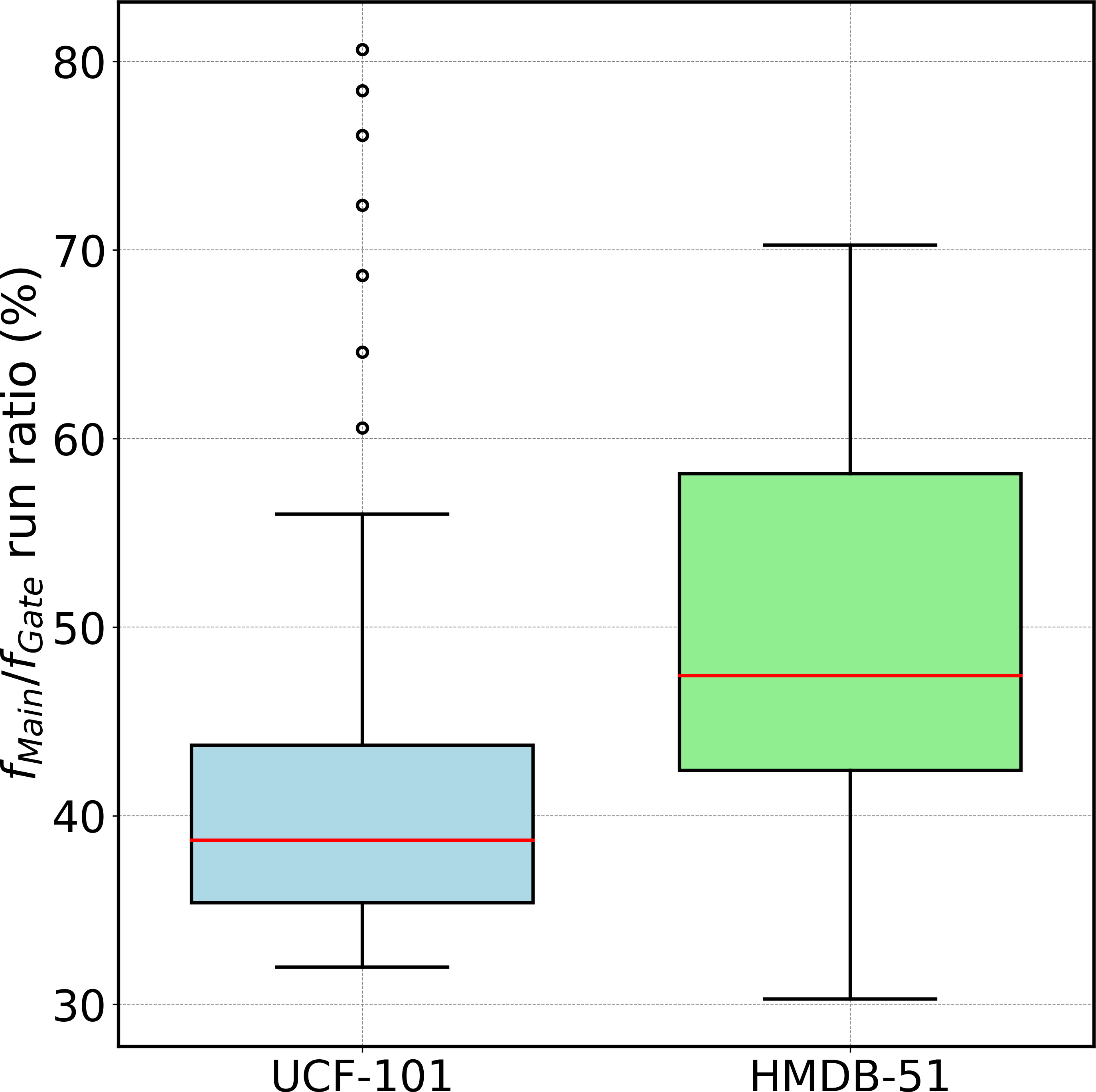}}%
\caption{\sysname early-exit performance statistics of each category on the two datasets  } 
\label{F.class_bluule_ucf101}
\end{figure}

\textbf{Performance on Categories.}
Figure \ref{F.class_bluule_ucf101} provides detailed early-exit performance statistics for \sysname across each category on the two datasets, including the number of frames viewed and the ratio of $f_{Main}$ activations. 
As shown in Figure \ref{F.class_bluule_ucf101}a, on the UCF101 dataset, \sysname requires fewer than 5 frames for 75\% of the categories to make a decision, while on the HMDB-51 dataset, this increases to 12 frames. This trade-off is justified by the improvement in accuracy. This demonstrates the effectiveness of our designed stop condition for gating module, which adaptively stops early during video played, avoiding unnecessary waiting.

As shown in Figure \ref{F.class_bluule_ucf101}b, for more than 75\% of the categories on both datasets, the activation ratio of $f_{Main}$ does not exceed 44\% and 68\%, respectively. This efficiency is due to the effective guidance provided by $f_{Gate}$. 
In contrast, traditional multimodal methods need inferring each modality of the sampled frames, resulting in the running time increasing rapidly as the product of the number of modalities and the number of sampled frames. 
This approach is both time-consuming and inefficient for online tasks.
\sysname employs a lightweight $f_{Gate}$ to initially predict each frame and determine whether to activate the corresponding $f_{Main}$ for further inference. This design allows \sysname to extract valuable information from each modality without incurring excessive computational overhead.


\begin{figure}[htb]
	\centering \includegraphics[width=1\columnwidth]{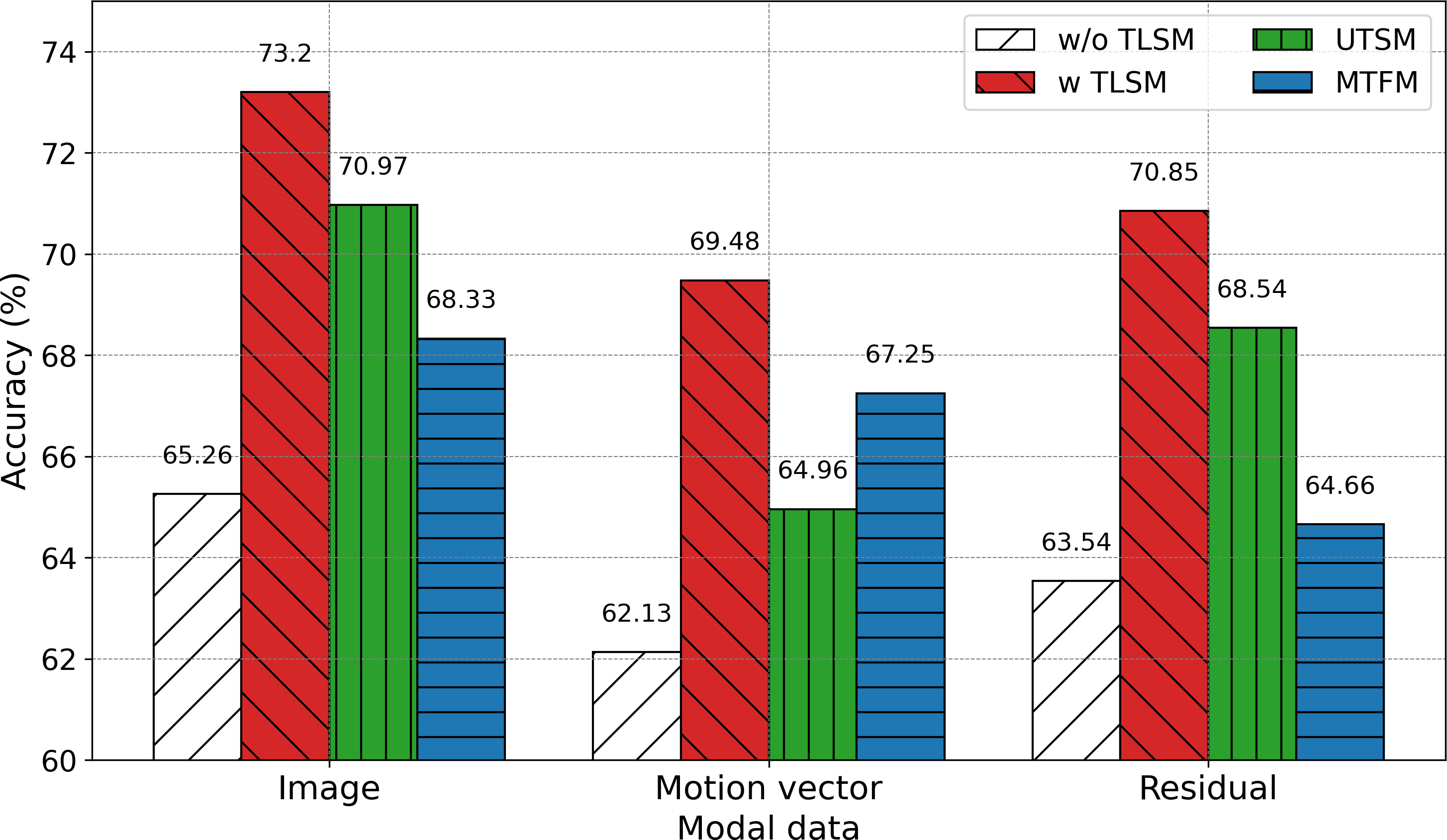}
	\caption{Performance Comparison of TLSM with UTSM and MTFM.
    }
	\label{F.TLSM}
\end{figure}

\textbf{TLSM.}
We compare TLSM with two shift modules used in action recognition: the Uni-directional Temporal Shift Module(UTSM) \cite{lin2019tsm} and the Motion Temporal Fusion Module(MTFM)\cite{guo2023lae}. The results are showed in Fig.\ref{F.TLSM}. For the classification task, TLSM incurs an additional latency of about 1 ms compared to the other two methods but achieves accuracy improvements of 3.14\% and 7.13\% on images. Additionally, it achieves at least a 3\% accuracy improvement on the other two modalities.
This improvement stems from the fact that simply shifting the previous frame is insufficient for continuous dense sampling in the online process. 
While TLSM allows exchanging frames that are temporally farther away with a larger field of view,  increasing accuracy by sharing temporal information effectively.

\textbf{MSEM.}
We evaluate the performance of the MSEM module in two key metrics: latency and accuracy. And compare it with other lightweight attention methods, Point-wise Spatial Attention (PSA) \cite{zhao2018psanet} and SimAM \cite{yang2021simam}. These attention mechanisms consume approximately 1-2 ms, confirming their lightweight nature. 
In terms of accuracy, the MSEM module achieves more than a 3\% improvement compared to the configuration without the attention enhancement module. Compared to PSA, the MSEM module increases accuracy by 2.73\%, 8.32\%, and 6.35\% on the three modalities, respectively. Compared to SimAM, the accuracy improvements are 3.56\%, 1.62\%, and 5.65\%, respectively.
This improvement arises because MSEM use additional MBs information, enabling \sysname to better focus on more complex spatial locations.

\begin{figure}[htb]
	\centering \includegraphics[width=1\columnwidth]{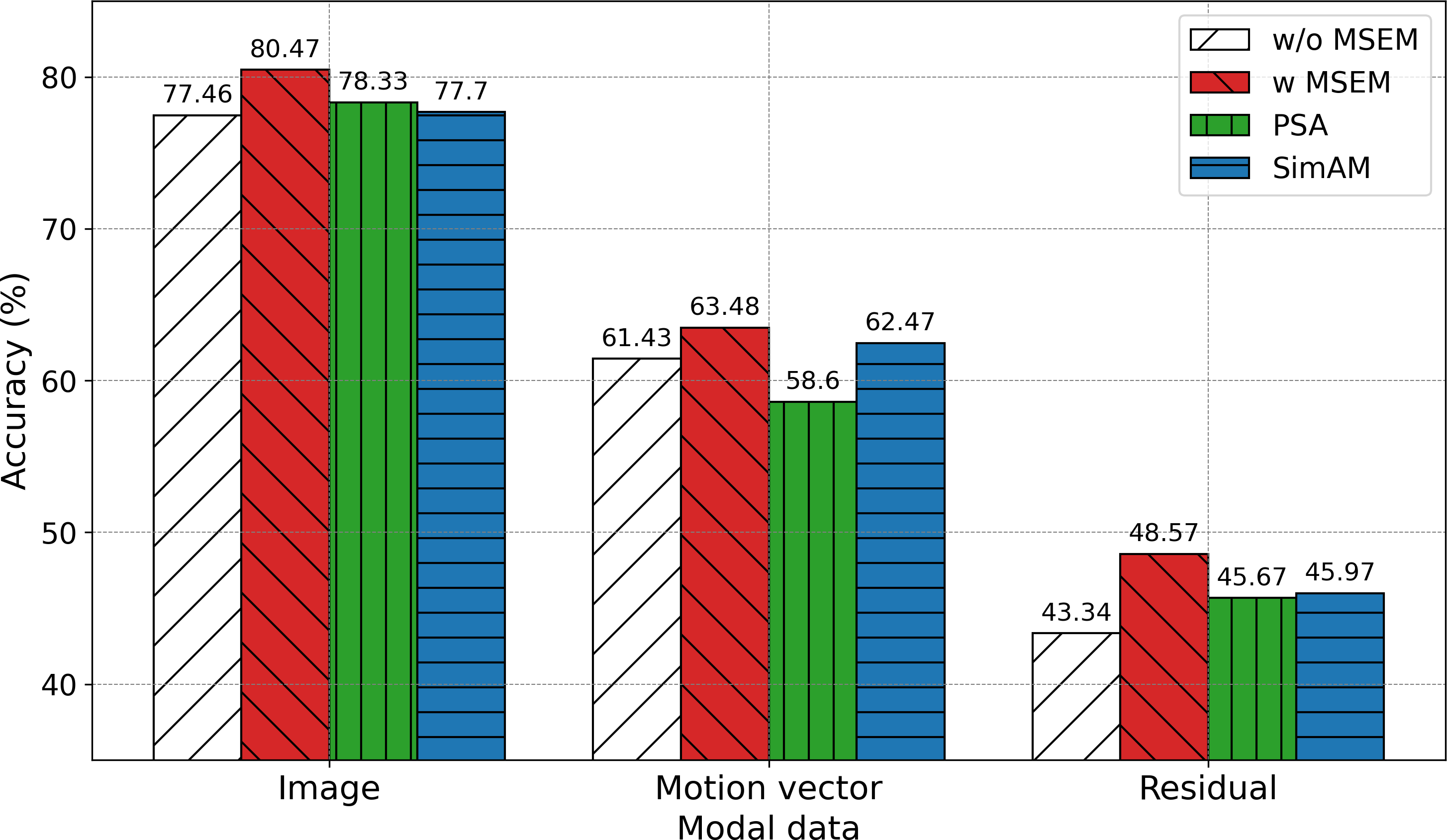}
	\caption{Performance Comparison of MSEM with SimAM\cite{yang2021simam} and  PSA\cite{zhao2018psanet}.
    }
	\label{F.MSEM3}
\end{figure}



\begin{figure}[htb]
\centering
\captionsetup[subfloat]{labelfont=small,textfont=normalsize,farskip=0pt}
\subfloat[Img]{
\includegraphics[width=0.49\columnwidth]{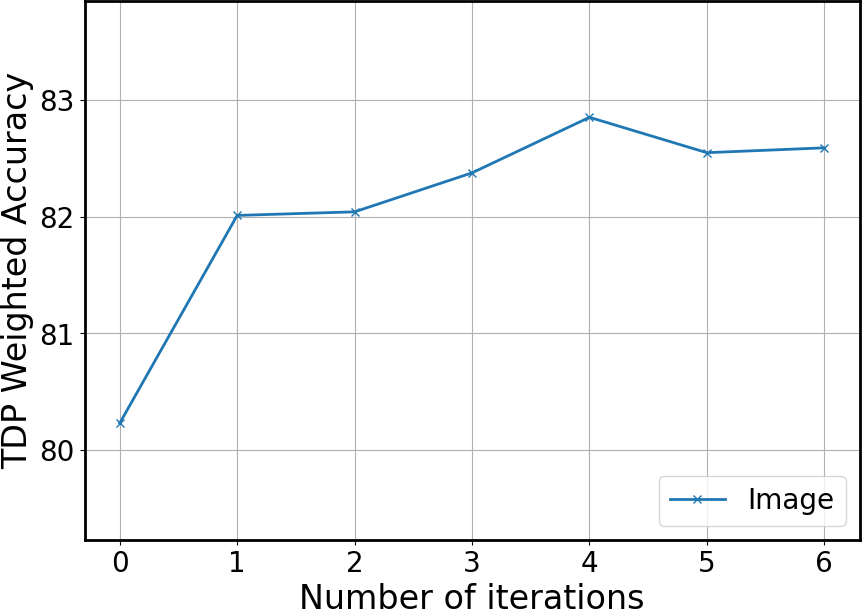}
}
\subfloat[MVs \& Res]{\includegraphics[width=0.49\columnwidth]{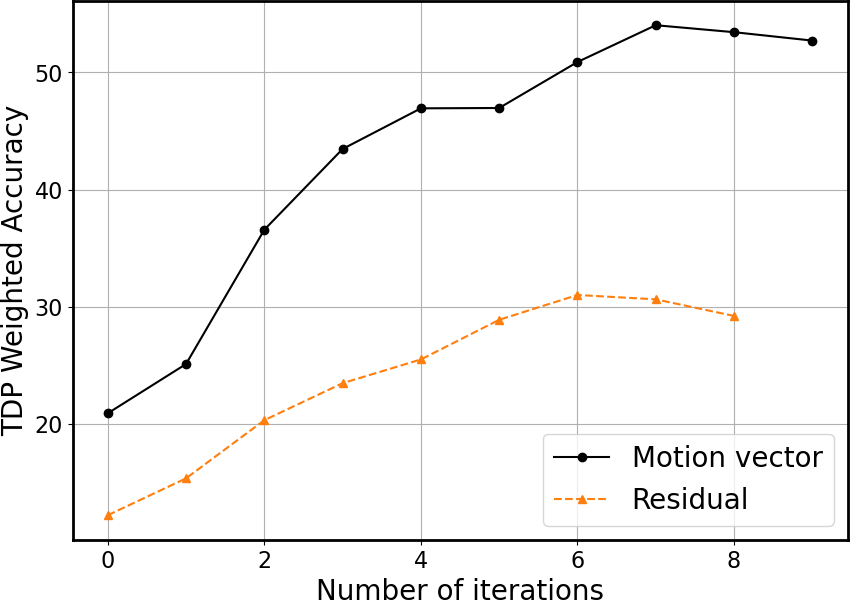}}%
\caption{The impact of iteration times on accuracy.
} 
\label{F.iteration_traing}
\end{figure}

\textbf{Iterative Training.}
We examine the correlation between the number of training iterations and accuracy, as shown in Fig.\ref{F.iteration_traing}. To better evaluate the early feature representation by \sysname, we use the weighted accuracy calculated according to Equ.\ref{E.TDP}. As the number of training iterations increases, the accuracy gradually improves. This enhancement is attributed to $f_{Gate}$ guiding the skipping of frames with incorrect or low-value inferences, allowing $f_{Main}$ to focus on frames with valuable and reliable information, thereby promoting a more effective learning process. This leads to continuous accuracy improvements as training progresses, enabling $f_{Main}$ to capture more useful features.

However, when a certain number of iterations is reached, the weighted accuracy plateaus and may slightly decline. This phenomenon can be attributed to several factors. First, overfitting of $f_{Gate}$. While it initially helps by excluding unhelpful frames, it may later mistakenly discard frames containing useful information. Second, overfitting of $f_{Main}$. As training iterations increase, $f_{Main}$ starts to learn noise and idiosyncrasies in the training set, negatively impacting its generalization ability.
Therefore, we select the weights at the peak of the weighted accuracy.

For different modalities, the accuracy increase for images is relatively small, whereas the accuracy of motion vectors and residuals improves by 157.68\% and 153.46\%, respectively, compared to no iterative training. This improvement occurs because their initial accuracy is lower. Through multiple iterations of training, $f_{Main}$ refines its attention based on $f_{Gate}$, and the guiding capability of $f_{Gate}$ is enhanced, effectively avoiding the inference of low-quality frames.

\textbf{IIE and MC-driven Fusion Module.}
We compare the performance of our proposed multi-modal feature fusion method with other fusion methods and conduct ablation experiments, as shown in Table.\ref{T.accuracy}.
In terms of accuracy, the fixed weighting scheme with the introduction of the IIE prior improves by 5.56\% compared to the average weighting scheme, demonstrating that IIE can better reflect the advantages of different modalities than the (1:1:1) weighting. Additionally, it approaches the performance of the TTP fusion method, indicating the effectiveness of the IIE prior information. However, it still falls short compared to learnable methods, suggesting that while prior knowledge enhances performance, it lacks the flexibility to capture valuable features from each modality.

After incorporating the IIE-driven fusion module, the accuracy with TDP weighting further improves by 15.28\%. Compared to the TPP and SCP fusion modules, the accuracy is enhanced by 14.00\% and 11.63\%, respectively. This improvement demonstrates that the IIE prior information enables better learning of dominant features among modalities. Adding the MC-driven fusion module further boosts accuracy by 6.12\%, indicating that differential fusion among modalities helps \sysname extract features more effectively, especially at the beginning of the video.

Regarding latency, we report the running latency on both CPU and GPU. The computational cost of the fixed weighting method is negligible, with latency close to zero. The CPU latency of the learnable method is very low, typically less than 1 ms. On the GPU, our method achieves a latency of 3.55 ms, representing a 68.76\% reduction compared to the TTP method and a 34.74\% reduction compared to the SCP method. The TTP method exhibits significant latency on the GPU due to its fusion of pooled row vectors, which does not leverage the GPU's matrix acceleration capabilities. Both the SCP and our fusion methods utilize square features at the tail of the model. Our method, by introducing prior information, reduces the computational load, resulting in lower GPU latency. This indicates that our algorithm optimizes latency while maintaining high accuracy, making it suitable for scenarios requiring quick responses.

\begin{table}[hbt]
	\setlength{\abovecaptionskip}{0pt}
\setlength{\belowcaptionskip}{1pt}
\renewcommand\arraystretch{1.2} 
	\caption{Performance comparison of different lightweight multi-modal feature fusion methods.}
	\label{T.accuracy}
	\centering
	\begin{threeparttable}
		\setlength{\tabcolsep}{1.8mm}{
			\begin{tabular}{lccc}
				\toprule[1.5pt] 
\multirow{2}{*}{Method}   & \multicolumn{1}{c}{TDP weighted} & \multicolumn{2}{c}{Lantency(ms)} \\
                             & accuracy(\%)      & CPU            & GPU             \\
\midrule[0.5pt]
Avg weighted\cite{wu2018compressed,guo2023lae}                & 70.67    & 0.01           & 0.01        \\
TPP fusion module   \cite{huo2020lightweight}                       & 74.88    & 0.33           & 11.32           \\
SCP fusion module  \cite{zheng2023dynamic}                  & 76.47    & 0.64           & 5.44            \\
\hdashline
IIE-weighted                         & 74.06    & 0.01           & 0.01                  \\
+IIE-driven fusion module     & 85.36    & 0.68           & 3.54           \\
+MC-driven fusion module     & 90.58    & 0.69           & 3.55             \\
\bottomrule[1.5pt]
		\end{tabular}}
	\end{threeparttable}
\end{table}

\section{Related Work}
\label{s.related_Work}

Deep learning-based action recognition leverages multiple networks to independently capture spatial and temporal data.
For instance, Two-Stream ConvNets\cite{simonyan2014two} employ a dual-network architecture to concurrently extract spatial and temporal information.
Building on this foundation, Temporal Segment Networks (TSN)\cite{wang2018temporal} balance temporal and spatial feature extraction by segmenting and aggregating video data. 
It is a flexible, expandable framework adaptable to diverse inputs.
Numerous works \cite{lin2019tsm, wang2021tdn, yang2022sta} based on the TSN framework enhance the capture of spatial and temporal features without significantly increasing computational cost by incorporating different attention modules and improving sampling methods. 
However, these methods rely on optical flow for motion representation, incurring substantial computational costs.
Simultaneous processing of two independent networks for spatial and temporal flow increases the demand for computing resources during training and inference.

With the develop of deep learning, some studies explore the use of backbones with stronger modeling capabilities to capture spatial and temporal features in image sequences. For example, 3D Convolutional Neural Networks (3D CNNs) \cite{liu2018t, dos2019cv} extend 2D convolution to the temporal dimension, enabling direct learning of spatio-temporal video features. Transformer models, on the other hand, leverage robust attention mechanisms to extract explicit spatial features and implicit temporal relationships between frames.
However, despite these improvements in accuracy, the high computational requirements due to the large number of parameters in these methods limit their applicability in real-time systems. For instance, the SOTA action recognition model, VideoMAE-V2 \cite{wang2023videomae}, processes only one frame every 5 seconds on the latest Snapdragon 8 Gen1+ SoC. This latency is unacceptable even for offline tasks.

Therefore, leveraging cost-effective and efficient compressed domain processing on edge devices is crucial for compensating the accuracy loss inherent in lightweight architectures.
Zhang et al.\cite{zhang2016real} design an enhanced motion vector CNN that serves as an alternative to  optical flow calculations, resulting in significant acceleration.
Coviar\cite{wu2018compressed} validate the feasibility and efficiency of training deep networks directly on compressed videos compared to raw video processing.
Slow I-frame Fast P-frame(SIFP) \cite{li2020slow} performs  inter-branch feature exchange to fuse multi-modal features.
CoViFocus\cite{zheng2023dynamic} first employs motion vectors for dynamic spatial focusing to identify task-related patches, followed by specific action classification on these patches.
However, this solution increase computational demands during inference.
LAE-Net \cite{guo2023lae} incorporates a motion temporal fusion module (MTFM) for enhanced motion information extraction and a dual compression knowledge distillation module (DCKD) for knowledge transfer from optical-flow-trained teacher networks to compressed-domain-trained student networks.
It achieves the highest accuracy among compressed domain action recognition methods with a small inference burden. A key challenge among these methods is the effective integration of information from dual streams.
Furthermore, inferring each modality of the sampled frames significantly increases the inference cost compared to image-only inference.
Additional studies focus on improving efficiency while maintaining high performance. 
FrameExit \cite{ghodrati2021frameexit} and TLEE \cite{wang2023tlee} perform early exits on sampled frames to reduce computational cost and inference latency. However, these works need to perform uniform sampling after the video playback ends, which still requires the whole video as a prior. 


\section{Conclusion and future}
\label{s.conclusion_future}

In this paper, we present \sysname, a framework designed  for online action recognition.  
\sysname initiates operation at the action's onset and can terminate early for yielding prediction results, eliminating the need to process the entire video segment.
To maintain accuracy while enabling early exit, we design two submodule, TLSM and MSEM  to bolster feature extraction capabilities of TFEM in both temporal and spatial dimensions.
We design an iterative training approach that alternately trains $f_{Gate}$ and $f_{Main}$, enabling \sysname to incrementally extract features from video initial segments.
Additionally, to boost classification accuracy, we harness data from multiple video modalities and integrate features across different modalities in both space and time using an IIE and MC-driven fusion module.
Our system achieves over 100$\times$ speed and power efficiency improvements compared to STOA action recognition solutions and outperforms hightest offline accuracy early exit methods in accuracy, meeting the application requirements for edge devices.

In the future, we aim to enhance the multi-modal feature fusion module by advancing the fusion timing, thereby further improving the features of non-image modalities.
Additionally, we will explore action recognition for multiple actions in online videos on edge devices, combining early exit and adaptive sliding windows to balance recognition accuracy and device energy consumption.



\bibliographystyle{IEEEtran}
\bibliography{references}
\end{document}